# Modeling and Analysis of Unmanned Remote Guided Vehicle on Rough and Loose Snow Terrain

Abhishek D. Patange[1], Sharad S. Mulik[2], R. Jegadeeshwaran[3], Dhananjay R. Jadhav[4], Prateek J. Ghatage[5], Gaurav R. Doshi[6], Rushikesh V Raykar[7]

[1]Assistant Professor, Department of Mechanical Engineering, College of Engineering Pune, Pune
[2]Professor, Department of Mechanical Engineering, RMD Sinhgad School of Engineering, Pune
[3]Associate Professor, School of Mechanical Engineering, Vellore Institute of Technology, Chennai
[4-7]Student, Department of Mechanical Engineering, Trinity Academy of Engineering, Pune

*Abstract*—Survival in remote snow bounded areas is unsafe and risky for mankind. Many problems like arthritis, frostbite, asthma, starvation can caused and lead to death. Indian Military provides transportation vehicles which are heavily built and needs manpower for monitoring. Hence it necessitates facilitating compact transportation to fulfill all requirements. This research aimed at design and analysis of mobile unmanned vehicle for transportation & providing medical help, food and other essential things necessary for surviving in such areas. This can also be used for military services to save the life of solider with less risk. It is typical medium weight, high speed vehicle which carries up to 35 kg load and can negotiate through loose snow, rough terrain with use of caterpillar track. The noteworthy feature of the vehicle is that it constitutes of spiral blades and V shape snowplow to make its way through snow. Hence it will repel the snow in outward direction for self-extraction. It also incorporates skis and hubs for changing the direction and smooth suspension. 3D model of the vehicle is drafted in CATIA and structural analysis is carried out in ANSYS. Control system design and mechatronics integration is proposed to develop the prototype by assembling various components.

I. INTRODUCTION

Landslide is a common feature in the snow bound regions of Himalayas, causing the loss of many precious lives of troops and civil population as well as property worth millions every year [1]. Survival in remote snow bounded areas is unsafe and risky for mankind. Many problems like arthritis, frostbite, asthma, starvation can caused and lead to death. Indian Military provides transportation vehicles which are bulky and needs manpower for monitoring. Many times it also got stuck in loose snow because of its heavy built. Hence it necessitates facilitating compact transportation to fulfill all requirements.

Here we describe design and analysis of mobile unmanned vehicle for transportation & providing medical help, food and other essential things necessary for surviving in such areas. This can also be used for military services to save the life of solider with less risk. It is typical medium weight, high speed vehicle which carries up to 35 kg load and can negotiate through loose snow, rough terrain with use of caterpillar track. The noteworthy feature of the vehicle is that it constitutes of spiral blades and V shape snowplow to make its way through snow. Hence it will repel the snow in outward direction for self-extraction. It also incorporates skis and hubs for changing the direction and smooth suspension. The sledge is provided on back side that helps the vehicle to float on snow. The skis are provided at bottom of vehicle to ensure steady position. It uses powerful DC motor of toque 80 Nmm with speed of 200 rpm. It have long range transmitter and receiver of 6 channel with range up to 1 km and control the motor with help of electronic speed controller (ESC). 3D model of the vehicle is drafted in CATIA and structural analysis is carried out in ANSYS. According to analysis and testing this vehicle is safe, cost effective, and compact. Control system design and mechatronics integration is proposed to develop the prototype by assembling various components.

II. LITERATURE REVIEW

Task-specific designs of unmanned ground vehicles for extreme climate have been developed over the time. Snow covered terrain poses many challenges for small and lightweight ground vehicles. Sinking in snow, resisting to compaction of snow, losing traction and snow entering into the driving system are a few to name [1-3]. The mobility of wheeled vehicles is limited by deep snow where the snow depth is greater than about twice the ground clearance of the vehicle. For mobility of tracked vehicles over deep snow, Mellor [4,5] suggested that the vehicle should have ground pressure less than about 7 kPa.

A ground vehicle with smaller ground clearance will encounter deep sinkage and will fail to gain support from the underlying snow surface. By designing a vehicle with low track pressure (<1.5 kPa), more support from the underlying snow cover can be obtained [6-8]. However, with such a design skid-steering becomes very difficult as the outward track tends to loose traction if the vehicle makes a turn. Better solutions to drive lightweight vehicles over deep snow are to use shorter tracks or use articulated chassis. Lightweight robotic vehicles can be designed for over snow mobility. Sinkage, resistance to snow compaction, loss of traction and ingestion of snow into the driving system are some of the challenges that an unmanned lightweight tracked vehicle faces in snowbound terrain.

R. K. Das [9] presented development of a lightweight and unmanned remotely operated vehicle (ROV) is conceptualized

and developed as a technological solution. In this study, design and features of this vehicle, named HimBot, are presented along with the results obtained from tests carried over snow at one of the field research stations of SASE. The outcome of this work will help in developing an optimized design of an ROV for over-snow mobility for a variety of applications.

Carnegie Mellon University, USA designed a gasoline driven vehicle called NOMAD [10] which was used to find and classify meteorites in Antarctica. It weighed approx. 725 kg and had a speed of about 0.5 m/s.

University of Kansas is developing robotic vehicles under a project called PRISM to utilize them for measurement of thickness of ice-sheets and establish bedrock conditions in Greenland and Antarctica [11, 12]. An International team of scientists led by Maho [13] at University of Strasbourg, France has developed a small remote controlled wheeled rover to study the behavior of penguins in Antarctica. The rover was fitted with an RFID device to read the heart-rate monitors, already fitted in the penguins.

Cold Regions Research and Engineering Laboratory (CRREL), USA has designed and developed an autonomous solar driven ground vehicle called, Cool Robot [14], for providing assistance during scientific expeditions in Antarctica and Greenland. The 61 kg vehicle was designed with solar panels fitted on its four sides and at top to maximize utilization of solar power. CRREL has also developed a low cost, GPS guided and battery-powered unmanned ground vehicle with an articulated chassis called Yeti [15]. It was designed to carry out ground penetrating radar surveys in Polar Regions.

### III. PROPOSED MODEL AND WORKING

To design mobile unmanned vehicle the model is described herewith. The mobile unmanned vehicle runs on caterpillar track mounted on three wheels. The main wheel rotates caterpillar track and motion is transmitted to corresponding two small sub-wheels which are assembled together. Small roller wheels are also provided in caterpillar track to prevent belt slip and ensure smooth suspension. The torque is generated at large wheel by motor which receives power from the battery. The speed of caterpillar track is regulated by means of Electronic Speed Controller (ESC).

Caterpillar track prevents skidding of the vehicle in loose snow. Loose and dense snow acts as an obstacle due to which vehicle may stuck. The skis are provided at bottom of vehicle to ensure steady position and smooth suspension of vehicle in snow. With the application of skis the direction of vehicle can also be changed. The sledge is provided on back side that helps the vehicle to float on snow. The vehicle can also badly get stuck in thick snow. In this condition the self-extraction can be done with the help of spiral blade by cutting hard snow. The crushed snow is repelled by the snowplow in outward direction. Snowplow is control by servo motor. The independent suspension system is provided to caterpillar track and skis to overcome bumps and to ensure smooth riding. The figure 1 shows block diagram of proposed vehicle.

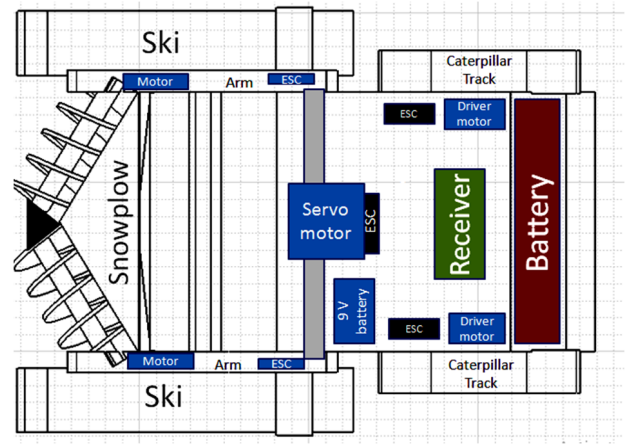

Figure 1: Block diagram of proposed vehicle

The system is simple, compact, cost effective and user friendly to carry high load with good speed. The Salient features of vehicle stated below.

- It facilitates smooth mobility in rough and loose snow terrain.
- It constitutes of spiral blades and V shape snowplow to make its way through snow. Hence it will repel the snow in outward direction for self-extraction.
- Load carrying capacity up to 35 kg, climb steep slope up to 65 degree, able to turn 360 degrees.
- The vehicle can negotiate through loose snow, rough terrain with use of caterpillar track. It also incorporates skis and hubs for changing the direction and smooth suspension.
- The sledge is provided on back side that helps the vehicle to float on snow. The skis are provided at bottom to ensure steady position. It uses powerful DC motor of toque 80 Nmm with speed of 200 rpm.
- It have long range transmitter and receiver of 6 channel with range up to 1 km and control the motor with help of electronic speed controller (ESC).

### IV. MODELING OF SYSTEM IN CATIA

The modeling of proposed system is carried out in CATIA. The figure 2 below shows 3D model of vehicle.

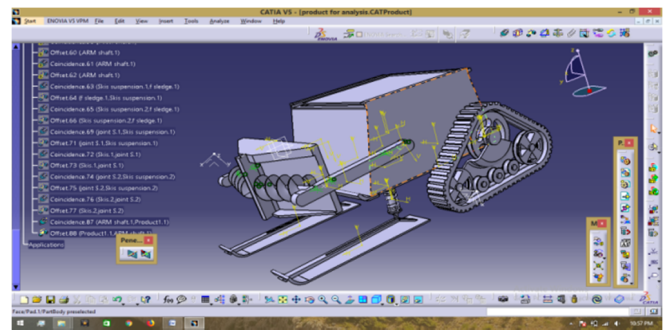

(a) 3D Model of proposed vehicle

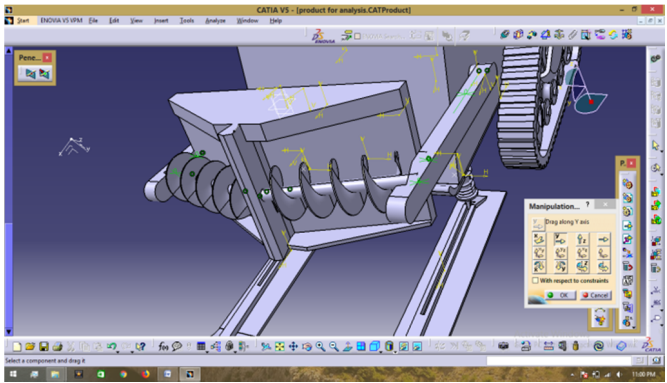

(b) Snowplow and spiral bldes

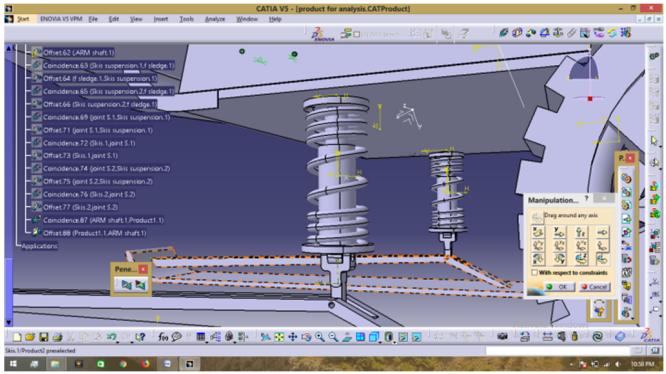

(c) Sring type suspension

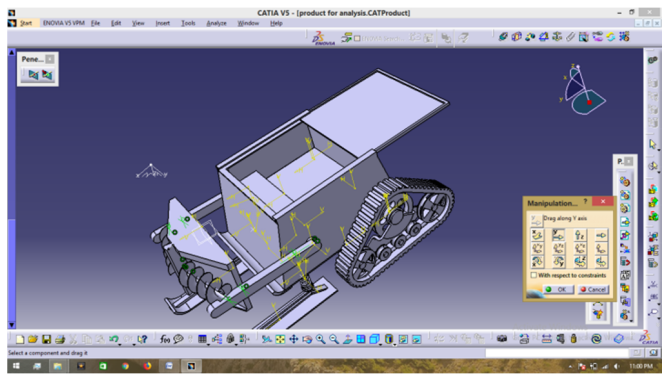

(d) Storage, caterpillar track, wheels

Figure 2: 3D Modeling of vehicle in CATIA

## V. ANALYSIS OF VEHICLE IN ANSYS

The static structural analysis of this vehicle is carried out in ANSYS.

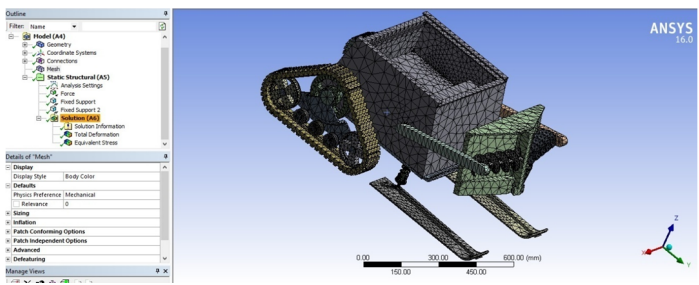

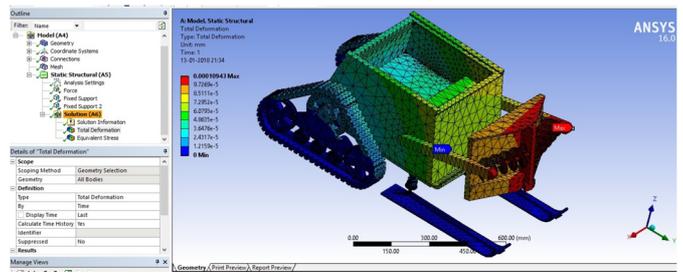

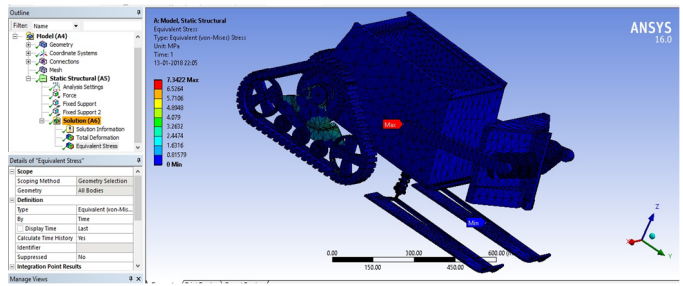

Figure 3: Analysis of vehicle carried out in ANSYS

The meshing is carried out, boundary conditions are applied and stress distribution has been found out from the analysis. The figure 3 shows the analysis of vehicle carried out in ANSYS.

## VI. CONTROL SYSTEM DESIGN & MECHATRONIC INTERFACE

A novel solution for development of proposed vehicle is formulated using control system design and mechatronic integration. It have two driver and two small motor for snow cutting. Driver motor is powerful motor due to high toque. It have servomotor of 35 kgcm which lift the snowplow and simultaneously blade also. All motors connected to ESC to control speed. One transmitter and receiver control all of it which used CB6T. Battery is used with maximum voltage which depends upon our vehicle running time. Figure 4 shows control system design and mechatronic integration.

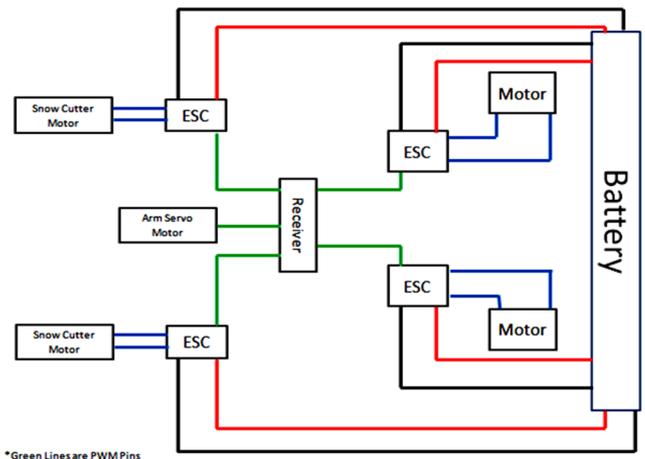

Figure 4: Control system design & mechatronic integration

The various components of proposed vehicle are:

- Li-ion battery
- Motor (Driver motor and Servomotor)
- Springs and Flanges
- Transmitter and receiver
- ESC (Electronics speed controller)
- Skis, Sledge
- Snowplow
- Spiral blades
- Caterpillar track

## VII. COST ESTIMATE FOR DEVELOPMENT OF MODEL

The cost of existing military vehicles are more than 20 Lakhs because of its heavy built. The overall cost of this vehicle is Rs. 1,13,800/-. The proposed system hence facilitates more features, is compact and of low cost. The cost estimate for development of this vehicle is given in table 1.

| SN | Name of component | Cost | Qty | Total Cost |
|----|---|---|---|---|
| 1 | DC Brush Gear Motor (300 rpm, 77 Nmm) | 21,240/- | 2 | 42,480/- |
| 2 | Servo Motor (35kgcm, Dual Shaft) | 5000/- | 1 | 5000/- |
| 3 | Caterpillar Track (3ft, Synthertic Rubber) | 9,940/- | 2 | 19,880/- |
| 4 | Skies | 1,500/- | 2 | 3000/- |
| 5 | Suspension (Motorcycle Spring Type) | 800/- | 4 | 3,200/- |
| 6 | Sled | 3,500/- | 1 | 3,500/- |
| 7 | Snow Cutting Motor | 5,000/- | 1 | 5,000/- |
| 8 | Transmitter & Receiver (Cb6t) | 3,000/- | 1 | 3,000/- |
| 9 | Battery | 6,030/- | 1 | 6,030/- |
| 10 | Control Unit | 4,000/- | 1 | 4,000/- |
| 11 | Materials | 9,910/- |  | 9,910/- |
| 12 | Machining | 8,800/- |  | 8,800/- |
|  | **Total** |  |  | **1,13,800/-** |

## VIII. CONCLUSION & FUTURE SCOPE

The design and analysis of mobile unmanned vehicle was carried out for transportation & providing medical help, food and other essential things necessary for surviving in such areas. It is typical medium weight, high speed vehicle which carries up to 35 kg load and can negotiate through loose snow, rough terrain with use of caterpillar track. The noteworthy feature of the vehicle is that it constitutes of spiral blades and V shape snowplow to make its way through snow. Hence it will repel the snow in outward direction for self-extraction. It also incorporates skis and hubs for changing the direction and smooth suspension. 3D model of the vehicle was drafted in CATIA and structural analysis was carried out in ANSYS. Control system design and mechatronics integration is proposed to develop the prototype by assembling various components. The cost of existing military vehicles are more than 20 Lakhs because of its heavy built but overall cost of proposed vehicle is Rs. 1,13,800/-. The proposed system hence facilitates more features, is compact and of low cost. The vehicles runs on DC power supply hence cost of fuels can be saved and scarcity of fuels in remote areas can be eliminated. This vehicle can be used for military services to save the life of solider with less risk.